\documentclass{IEEEcsmag}

\usepackage[colorlinks,urlcolor=blue,linkcolor=blue,citecolor=blue]{hyperref}
\usepackage{cite}
\usepackage{upmath}
\usepackage{amsfonts} 
\usepackage{algorithm}
\usepackage{algpseudocode}
\usepackage{amsmath,amssymb,amsfonts}
\usepackage{rotating}

\jvol{XX}
\jnum{XX}
\paper{8}
\jmonth{}
\jname{}
\pubyear{2022}

\begin{document}

\title{Fast Deep Autoencoder for Federated learning}

\author{David Novoa-Paradela}
\affil{CITIC, University of A Coruña, Spain; david.novoa@udc.es}

\author{Oscar Fontenla-Romero}
\affil{CITIC, University of A Coruña, Spain; oscar.fontenla@udc.es}

\author{Bertha Guijarro-Berdiñas}
\affil{CITIC, University of A Coruña, Spain; berta.guijarro@udc.es}

\markboth{Preprint version}{Fast Deep Autoencoder for Federated learning}

\begin{abstract}

This paper presents a novel, fast and privacy preserving implementation of deep autoencoders. DAEF (Deep Autoencoder for Federated learning), unlike traditional neural networks, trains a deep autoencoder network in a non-iterative way, which drastically reduces its training time. Its training can be carried out in a distributed way (several partitions of the dataset in parallel) and incrementally (aggregation of partial models), and due to its mathematical formulation, the data that is exchanged  does not endanger the privacy of the users. This makes DAEF a valid method for edge computing and federated learning scenarios. The method has been evaluated and compared to traditional (iterative) deep autoencoders using seven real anomaly detection datasets, and their performance have been shown to be similar despite DAEF's faster training.

\end{abstract}

\maketitle

\chapterinitial{As happened} at the time with the massive adoption of personal computers, the technological development of recent years has caused a substantial increase in the number of small computing machines such as smartphones or Internet of Things (IoT) devices, for both industrial and personal use. Despite their size, these devices have enough computing power to perform tasks that until a few years ago were considered unapproachable, such as the training of small machine learning models, real-time inference or the exchange of large amounts of information at high speeds. 

Due to the abundance of these devices and the inefficiencies of traditional cloud computing for applications that demand low latencies, a new computing paradigm called edge computing has emerged \cite{KHAN2019219}. Edge computing moves computing away from data centers to the edge of the network, bringing cloud computing services and utilities closer to the end user and their devices. This allows faster information processing and response time, as well as freeing up the network bandwidth.

From a machine learning point of view, this new technological scenario is very suitable for the application of federated learning \cite{XIA2021100008}. Federated learning is a collaborative machine learning scheme that allows heterogeneous devices with different private data sets to work together to train a global model. In addition to this collaborative learning, this work scheme emphasizes the preservation of the privacy of local data collected on each device by implementing mechanisms that prevent possible direct and indirect leaks of their data. 

On the other hand, in machine learning, anomaly detection is the branch that builds models capable of differentiating between normal and anomalous data \cite{10.1145/1541880.1541882}. A priori, this turns anomaly detection into a classification problem with only two classes. However, since anomalies tend to occur sporadically, normal data are the ones that prevail in these scenarios, so it is common that models must be trained with only normal data.
The objective is to learn to represent the normal class with high precision in order to be able to classify new data as either normal or abnormal. In many real systems, the response time to a detection of an anomaly (failure) can be critical, as is the case with autonomous vehicles \cite{8744265} or industrial systems \cite{9139976}. The development of anomaly detection techniques based on edge computing and federated learning may be the solution to reduce these response times.  

In this paper we introduce  DAEF (Deep Autoencoder for Federated learning), a fast and privacy‐preserving deep autoencoder for edge computing and federated learning scenarios. Unlike traditional deep neural networks, its learning method is non-iterative, which drastically reduces training time. Its training can be carried out in a distributed way (several partitions of the dataset in parallel) and incrementally (aggregation of partial models), and due to its mathematical formulation the data that is exchanged  does not endanger the privacy of the users. All of this makes DAEF a valid method for edge computing and federated training scenarios, capable of performing tasks as anomaly detection on large datasets while maintaining the performance of traditional (iterative) autoencoders. 

This document is structured as follows. Section 2 contains a brief review of the main anomaly detection techniques for edge computing, providing an overview of this field. Section 3 describes the ideas taken as the basis for the development of the proposed DAEF method and Section 4 describes its operation. Section 5 discusses DAEF's privacy-preserving capabilities. Section 6 illustrates the performance of DAEF through a comparative study with traditional autoencoders. Finally, conclusions are drawn in Section 7.

\section{2. RELATED WORK}\label{relatedwork}

Anomaly detection is a field that has a large number of algorithms that solve the problem of distinguishing between normal and anomalous instances in a wide variety of ways \cite{chandola, khan}. Depending on the assumptions and processes they employ, in traditional anomaly detection we can distinguish between five main types of methods : probabilistic, distance-based, information theory-based, boundary-based, and reconstruction-based methods. In general, these algorithms are characterized by their high performance when classifying new data, however they do not focus on other aspects which from a centralized perspective may seem less important, such as data privacy and incremental learning. This makes it difficult to apply many of these classical methods in decentralized environments. For this reason, the strong expansion of edge computing has brought with it a new line of research in the field of anomaly detection in charge of designing new algorithms capable of learning in a distributed and, in some cases, incremental way, while preserving data privacy. Due to their good performance, it is common for these methods to be based on reconstruction (neural networks). In this section we will distinguish between reconstruction based methods that use autoencoders \cite{10.5555/1756006.1953039} and those that do not.

Among those that do not use autoencoders is DÏOT \cite{nguyen2019diot}, a self-learning distributed system for security monitoring of IoT devices which utilizes a novel anomaly detection approach based on representing network packets as symbols, allowing to use a language analysis technique to detect anomalies. B. Hussain \textit{et al}. \cite{8844663} presented a deep learning framework to monitor user activities of multiple cells and thus detect anomalies using feedforward deep neural networks. R. Abdel \textit{et al}. \cite{sater2021federated} introduced a federated stacked long short-time memory model to solve multi-task problems using IoT sensors in smart buildings. Y. Zhao \textit{et al}. \cite{10.1145/3368926.3369705} propose a multi-task deep neural network in federated learning to perform simultaneously network anomaly detection, VPN traffic recognition, and traffic classification. Other authors like D. Preuveneers \textit{et al}. \cite{app8122663} propose the use of blockchain technology to carry out a decentralized registry of federated model updates. This guarantees the integrity of incrementally-learned machine learning models by cryptographically chaining one machine learning model to the next. These solutions obtain good results, however they do not emphasize privacy preservation and their iterative learning can lead to long training times.

On the other hand, if we focus on autoencoders \cite{10.5555/1756006.1953039}, it is also possible to find works oriented towards edge computing and/or federated learning scenarios. Autoencoders (AE) are a type of self-associative neural network whose output layer seeks to reproduce the data presented to the input layer after having gone through a dimensional compression phase. In this way, they manage to obtain a representation of the input data in a space with a dimension smaller than the original, learning a compact representation of the data, retaining the important information and compressing the redundant one. For this reason, they are widely used for the elaboration of models that are robust to noise, an important quality in anomaly detection and regression problems. {\bf Figure 1} represents the traditional architecture of an autoencoder network. 

\begin{figure}
\centerline{\includegraphics[width=18.5pc]{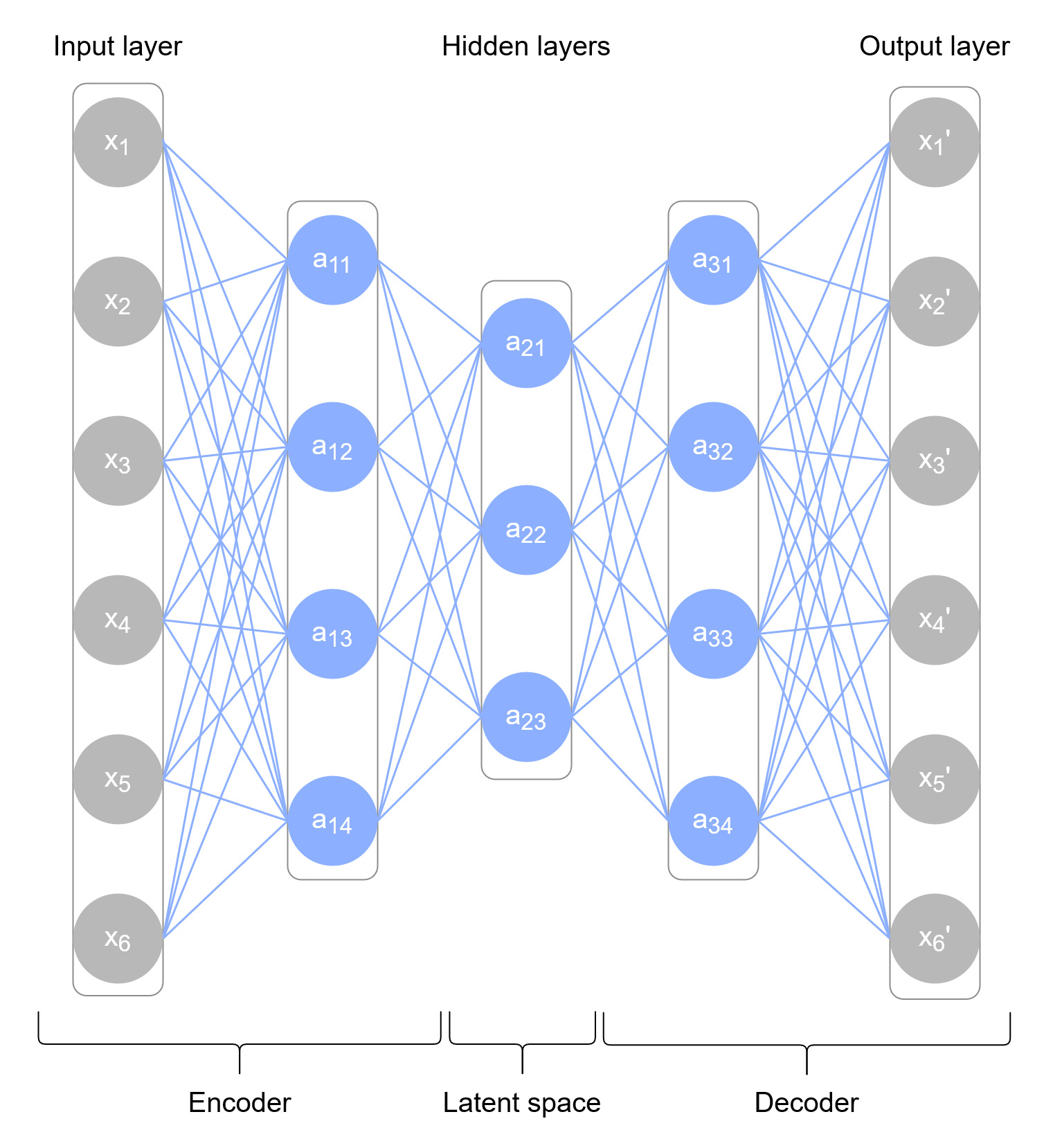}}
\caption{Example of autoencoder neural network architecture.}
\label{figureAE}
\end{figure}

T. Luo \textit{et al}. \cite{8422402} propose to use autoencoders for anomaly detection in wireless sensor networks, however each edge device does not train a local model with its own data. These devices send their local data to a central cloud node from which the training of the global model is carried out. In the approach presented by M. Ngo \textit{et al}. \cite{ngo2020adaptive}, an adaptive hierarchical edge computing system composed by three autoencoder models of increasing complexity is used for IoT anomaly detection. 

In the two previous works, as well as in the majority that use this type of networks, the autoencoders are trained during several iterations to adjust their parameters (weights, bias) using techniques such as the gradient descent and backpropagation. This greatly increases training time, specially when dealing with large datasets or complex networks architectures, which in edge computing scenarios can be critical. 

However, there is a line of work that allows training autoencoders in a non-iterative way. This is based on Extreme Learning Machines (ELM) \cite{HUANG2006489}, an alternative learning algorithm originally formulated  for single-hidden layer feedforward neural networks (SLFNs). This algorithm tends to provide good generalization performance and an extremely fast learning speed. Over time, more advanced versions such as MLELM \cite{KasunRepresentationalLW}, a multilayer version of ELM, or DELM \cite{DELM}, a deep version of ELM, have been developed.

For anomaly detection in edge computing and federated learning scenarios, R. Ito \textit{et al}. \cite{2021SSSS} propose to combine OS-ELM (Online Sequential Extreme Learning Machine) \cite{4012031} with autoencoders. This allows each edge device to train its own local model and incrementally update it with the results obtained by the other devices. Nevertheless, a possible limitation of this solution is its autoencoder architecture with only one hidden layer, which in some cases may not be sufficient.

In this work we present DAEF, a deep autoencoder with the following characteristics:

\begin{itemize}
    \item The architecture is deep and asymmetrical. 
    \item The training process is non-iterative.
    \item It can be trained in a distributed and incremental way.
    \item It is a privacy-preserving method.
\end{itemize}

\section{3. BACKGROUND} \label{background}

This section introduces the theoretical foundations of the three methods taken as the basis for the development of DAEF: (a) DSVD-autoencoder \cite{https://doi.org/10.1002/int.22296}, a Distributed and privacy-preserving autoencoder for anomaly detection using Singular Value Descomposition; (b) MLELM \cite{KasunRepresentationalLW}, a Multilayer Extreme Learning Machine implementation with a layer-by-layer training process; (c) ROLANN \cite{DBLP:conf/iwann/Fontenla-Romero21}, a novel Regularized training method for One-Layer Neural Networks. 

\subsection{3.1 Distributed Singular Value Decomposition Autoencoder}\label{3_1}

DSVD-autoencoder (Distributed Singular Value Decomposition-Autoencoder) \cite{https://doi.org/10.1002/int.22296} is a hidden single-layer autoencoder network for anomaly detection. The aim in the encoder is to learn a vector space embedding of the input data extracting a meaningful but lower dimensional representation. To achieve this dimensionality reduction, the Singular Value Decomposition (SVD) of matrices is used. In the decoder, the goal is to reconstruct the input from the low-dimensional representation, in this case using LANN-SVD \cite{lannsvd}. The privacy-preserving properties, parallelization, and non-iterative training of this method make it a suitable alternative for anomaly detection in edge computing scenarios and a good basis for out work, although it has the limitation of only allowing the use of one hidden layer.

\subsection{3.2 Multilayer Extreme Learning Machine}\label{3_2}
MLELM (Multilayer Extreme Learning Machine) \cite{KasunRepresentationalLW} is a multilayer neural network that makes use of unsupervised learning to train the parameters in each layer, eliminating the need to fine-tuning the network. The novelty of this work is that it trains each layer by using an ELM-AE (Extreme Learning Machine-Autoencoder) \cite{KasunRepresentationalLW}, which is an unsupervised single hidden layer neural network that, like any autoencoder, tries to reproduce the input signal at the output. As a result, the authors obtain a mechanism to train deep networks in a non-iterative, fast, and mathematically simple way. This mechanism has served as an inspiration for the work presented here.

\subsection{3.3 Regularized One-Layer Neural Network}\label{3_3}
ROLANN (Regularized One-Layer Neural Networks) \cite{DBLP:conf/iwann/Fontenla-Romero21} is a training regularized by the L2 norm that allows to train single layer neural networks (without hidden layers) in a non-iterative, incremental, and distributed way while also preserving privacy. To do this, the method minimizes the mean squared error (MSE) measured before the activation function of the output neurons, as described in \cite{10.1016/j.patcog.2009.11.024}. The algorithm can be used incrementally and distributed, making it a perfect fit for federated learning environments.

\section{4. THE PROPOSED METHOD} \label{method}

The main objective of the proposed  method (Deep Autoencoder for Federated learning) is to learn a compressed representation of the normal data and to reconstruct the inputs to the output of the autoencoder from this reduced space. These tasks should be carried out in a distributed way, and incrementally where possible, in order to apply the algorithm in edge computing and federated learning environments. To achieve this, DAEF employs an asymmetric autoencoder architecture as shown in {\bf Figure 2}. A first single-layer \textit{encoder} reduces the dimensionality of the input data and it is adjusted using a distributed SVD process. It is followed by a multi-layer \textit{decoder} to reconstruct the input signal at the output which is trained in layer-by-layer basis through a non-iterative process. This section presents in detail the steps followed by the method and its theoretical foundations.

\begin{figure*}
\centerline{\includegraphics[width=38.5pc]{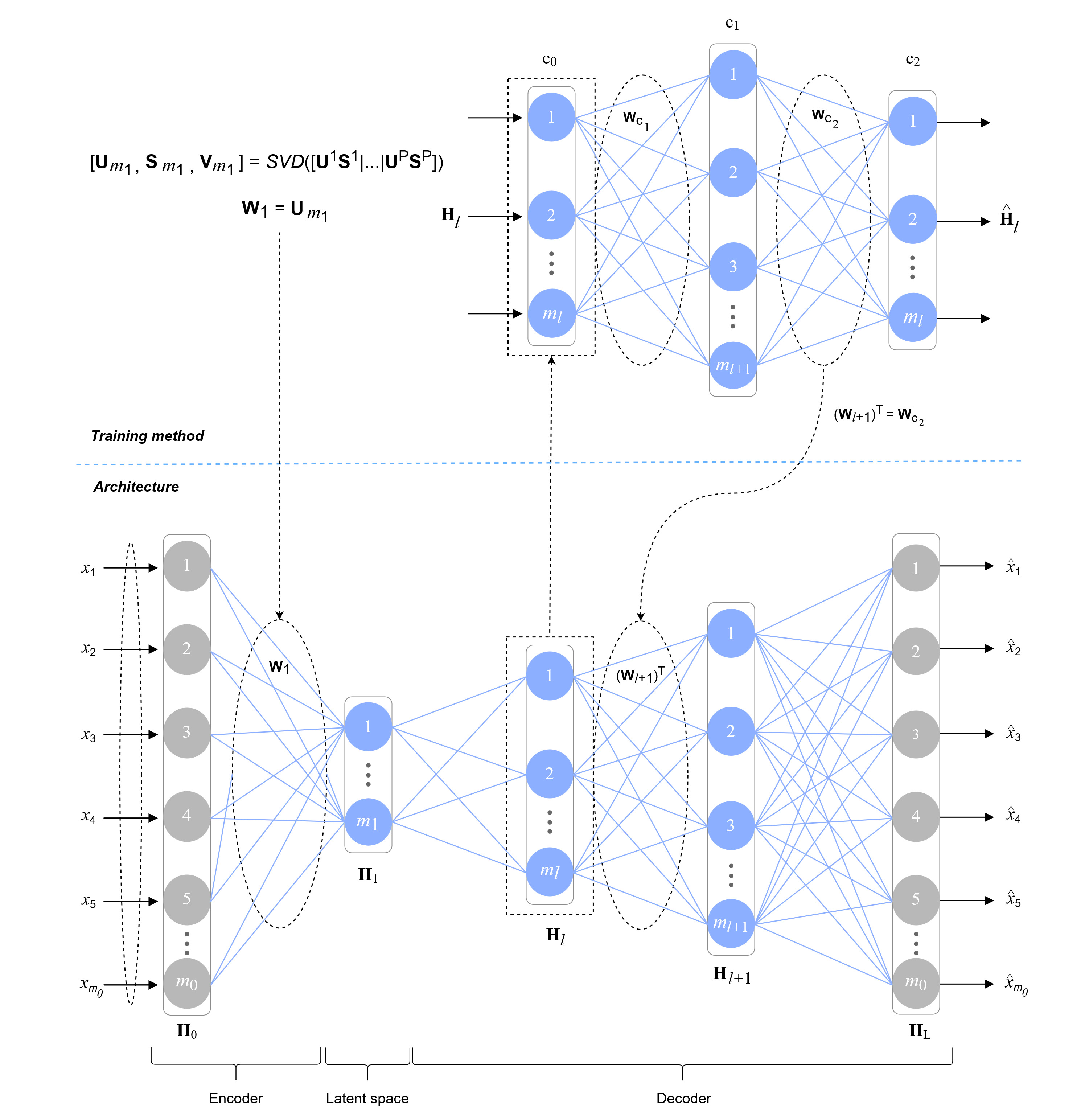}}
\caption{The asymmetric deep autoencoder DAEF.}
\label{arquitect}
\label{figureAEGenerica}
\end{figure*}

\subsection{4.1 The encoder}\label{4_1}

In the encoder, the goal is to learn a vector space embedding of the input data extracting a useful but lower‐dimensional representation, known as the latent space. This can be accomplished by a low‐rank matrix approximation, which is a minimization problem that tries to approximate a given matrix of data by another one subject to the constraint that the approximating matrix has reduced rank \cite{RePEc:spr:psycho:v:1:y:1936:i:3:p:211-218}. Given that the dimension of this new space is determined by the number of neurons $m_1$ of the first hidden layer, the rank-$m_1$ SVD of the input matrix $\mathbf{X}$ is used to obtain the weights \textbf{W}$_1$ of this first layer. 

The full SVD of $\mathbf{X} \in \mathbb{R}^{m_0\times n}$, where $m_0$ is the number of input variables and $n$ the number of data samples, is a factorization of the form:

\begin{equation}
\textbf{X = USV}^{T},
\end{equation}

\noindent where $\mathbf{S}\in\mathbb{R}^{m_0\times n}$ is a diagonal matrix with descending ordered non-negative values on the diagonal that are the singular values of \textbf{X}, while $\textbf{U}\in\mathbb{R}^{m_0\times m_0}$ and $\textbf{V}\in\mathbb{R}^{n\times n}$ are orthogonal matrices containing the left and right singular vectors of \textbf{X}. In a low‐rank approximation, the optimal rank‐$m_1$ approximation of \textbf{X} can be computed by taking the first $m_1$ columns of \textbf{U} and rows of \textbf{V}$^{T}$ and truncating \textbf{S} to the first $m_1$ diagonal elements. The new truncated matrices \textbf{U}$_{m_1}\in \mathbb{R}^{m_0 \times  m_1}$ and \textbf{V}$_{m_1}^{T}\in \mathbb{R}^{m_1 \times  n}$ are, respectively, $m_1$‐dimensional representations of rows (features) and columns (samples) of the input data \textbf{X}. Therefore, \textbf{U}$_{m_1}$ is used as the weights for the first layer \textbf{W}$_1 \in \mathbb{R}^{m_0 \times  m_1}$ as it contains the $m_1$‐dimensional transformation of the input space ($\mathbb{R}^{m_0}\rightarrow \mathbb{R}^{m_1}$). 

In a distributed scenario, the data matrix \textbf{X} is partitioned into $P$ several blocks, that is \textbf{X} = $\left [ \textbf{X}^1 | \textbf{X}^2 |\cdots  |\textbf{X}^P\right ]$. In this case, the SVD of the entire \textbf{X} can be also computed distributively (DSVD) by calculating at each site $p$ the local SVD ($\textbf{U}^p$ and $\textbf{S}^p$), corresponding to $\textbf{X}^{p}$, and then arbitrarily computing the following operation at any of the nodes \cite{2016dsvddistributed}:

\begin{equation}
\resizebox{.88\hsize}{!}{$[\textbf{U}_{m_1}, \textbf{S}_{m_1}, \textbf{V}_{m_1}] = SV\!D([\textbf{U}^1\textbf{S}^1 | \ldots | \textbf{U}^P\textbf{S}^P]).$}
\end{equation}

Therefore, the weights of the first layer $\textbf{W}_{1} = \textbf{U}_{m_1}$ are obtained collaboratively across all nodes locations. Finally, the outputs of the first hidden layer of the network can be calculated, at each location, as: 
\begin{equation}
\textbf{H}_{1}^p = f_{1}\left (  \textbf{W}_{1}^T \textbf{X}^{p} \right )\!; \forall \;   p = 1, ..., P,
\label{eq1}
\end{equation}

where $f_1$ is the activation function of the first hidden layer.

This dimensionality reduction method has been used, despite the existence of other techniques such as PCA (Principal Component Analysis), because, as has been demonstrated \cite{2016dsvddistributed}\cite{https://doi.org/10.1002/int.22296}, the distributed implementation of SVD performs well and preserves data privacy, which is very suitable for edge computing environments. It has been decided to use a single-layer encoder, that is, a single dimensionality reduction process using SVD, because chaining several SVD processes sequentially and progressively (one per layer) did not show better performance. 
    
\subsection{4.2 The decoder}\label{4_2}
    
In the decoder, the goal is to reconstruct the input from the low-dimensional representation provided by the output of the first hidden layer (see Equation (\ref{eq1})). In order to be able to work with large datasets in a fast and efficient way, we propose to apply a non-iterative learning method to obtain the decoder parameters. 

Similar to ELM-AE \cite{KasunRepresentationalLW}, DAEF employs an auxiliary network to determine the parameters of each layer of the decoder in an unsupervised way, layer by layer. In the DAEF decoder, the weights and bias of the $(l+1)$-th hidden layers will be calculated with an auxiliary network, which will use $f_{l+1}$ as activation function. The output matrix of $(l+1)$-th layer (\textbf{H}$_{l+1}$) is obtained as follows:
\begin{equation}
\textbf{H}_{l+1} = f_{l+1}\left ( \textbf{W}_{l+1}^T \textbf{H}_{l} +  \textbf{b}_{l+1} \textbf{1}^T\right )
\label{eqHi}
\end{equation}

\noindent being $\textbf{H}_{l}$ the output matrix of the $l$-th layer, $\textbf{W}_{l+1} \in \mathbb{R}^{m_{l}\times m_{l+1}}$ and $\textbf{b}_{l+1} \in \mathbb{R}^{m_{l+1}\times 1}$ the estimated weight matrix and bias vector of the layer, respectively, and $\textbf{1}$ a column vector of $n$ ones.

 The use of this auxiliary network is shown in {\bf Figure 2}, where \textbf{W}$_{l+1}$ represents the output weights of the auxiliary network and $m_l$ the number of neurons in a layer $l$. As can be seen, the auxiliary network is a single-hidden layer sparse autoencoder. To calculate the parameters between the $l$-th and the $(l+1)$ hidden layers, the number of neurons in the input and the output layers of the auxiliar network will be identical to $m_l$, and the number of neurons of his hidden layer will be $m_{l+1}$.

The training of this auxiliary network can be divided into two stages: the training of the first half of the network (layers c$_0$-c$_1$) in which the input received by the first layer (c$_0$) is transformed; and a second stage (layers c$_1$-c$_2$) in which, using the data coming from c$_1$ (\textbf{H}$_{c_1}$), the original input is reconstructed at the output of the network (\textbf{H}$_{c_2}$). 

The weights of the first stage (\textbf{W}$_{c_1}$) are fixed and obtained using the Xavier Glorot initialization scheme, while the bias vector ($\textbf{b}_{c_1}$) is randomly established using a normal distribution with zero mean and standard deviation equal to 1. Given this, the \textbf{H}$_{c_1}$ output of the hidden layer can be calculated as:
\begin{equation}
\textbf{H}_{c_1} = f_{c_1} \left ( \textbf{W}^T_{c_1}\textbf{H}_{c_0} + \textbf{b}_{c_1} \textbf{1}^T \right ),
\label{eqHc2}
\end{equation}

\noindent where $f_{c_1}$ is the activation function, \textbf{W}$_{c_1}$ are the fixed weights, $b_{c_1}$ the random bias, and \textbf{H}$_{c_0}$ the already known output of the DAEF's  $l$-th layer.

In the second stage, the weights \textbf{W}$_{c_2}$ are computed in a supervised way using the regularized ROLANN method \cite{DBLP:conf/iwann/Fontenla-Romero21} as:

\begin{equation}
\left [ \mathbf{U}^{p}, \mathbf{S}^{p}, \sim \right ] = SV\!D\left ( \mathbf{X}^p \mathbf{F}^p \right ),
\label{eqROLAN}
\end{equation}

\begin{equation}
 \mathbf{M}^{p} =  \mathbf{X}^{p} \ast \left ( \mathbf{f}^p. \ast \mathbf{f}^p. \ast \mathbf{\bar{d}}^p \right ),
 \label{eqROLANN}
\end{equation}

\begin{equation}
\left [ \mathbf{U}^{k \mid p}, \mathbf{S}^{k \mid p}, \sim \right ] = SV\!D\left ( \mathbf{U}^k \mathbf{S}^k \mid \mathbf{U}^p \mathbf{S}^p \right ),
\label{eqROLAN}
\end{equation}

\begin{equation}
 \mathbf{M}^{k \mid p} =  \mathbf{M}^{k} + \mathbf{M}^{p},
 \label{suma}
\end{equation}

\begin{equation}
\resizebox{.88\hsize}{!}{$\mathbf{W_{c_2}}  =  \mathbf{U}^{k \mid p} \ast inv\left ( \mathbf{S}^{k \mid p} \ast \mathbf{S}^{k \mid p} + \lambda \mathbf{I} \right ) \ast \left ( {\mathbf{U}^{k \mid p}}^{T} \ast \mathbf{M}^{k \mid p} \right ),$}
\end{equation}

where $\mathbf{\bar{d}}^p$ and $\mathbf{f}^p$ are the inverse and derivative of the neural function, respectively, at each data point, and $\mathbf{F}^p$ is the diagonal matrix of $\mathbf{f}^p$. $\mathbf{M}^{p}$, $\mathbf{U}^{p}$ and $\mathbf{S}^{p}$ correspond to the knowledge obtained in the $p$ partition, while $\mathbf{M}^{k}$, $\mathbf{U}^{k}$ and $\mathbf{S}^{k}$ correspond to the knowledge accumulated after several iterations of incremental learning.

Considering that each output of the neural network depends solely on a set of independent weights, this second stage can be computed in parallel if the device has several cores. Once this is calculated, the weights between the   DAEF's  $l$-th and $(l+1)$ layers can be obtained as $\textbf{W}_{l+1} = \textbf{W}_{c_2}^T$, and the output \textbf{H}$_{l+1}$ can be calculated using Equation (\ref{eqHi}). 

This process will be repeated for each of the hidden layers of the decoder, layer by layer, using the outputs of each one to calculate the weights of the next one, until reaching the last   DAEF's  layer. 

Finally, the output target values for the   DAEF's  last layer are known (the same as in the DAEF's  input layer), therefore the weights of the last layer can be calculated directly in a supervised and distributed way using ROLANN. The activation function for the last layer will be linear as we want to reconstruct the input data of the network (any real value) at the output. 

We can summarize the DAEF training as follows:

\begin{enumerate}
    \item Dimensionality reduction in the first layer using distributed SVD (encoder).
    \item Unsupervised/supervised training, layer by layer, using an auxiliary network in which ROLANN is used (decoder).
    \item Supervised training of the last layer using ROLANN method (decoder).
\end{enumerate}

\subsection{4.3 Incremental and distributed learning}\label{4_3}

DAEF performs various operations that can be computed in a distributed way if the node (device) on which it is executed has several cores. These operations are the SVD computation of the encoder (the dataset can be divided and the partial SVDs concatenated and recalculated) and the ROLANN regularization processes in the decoder (the weights with respect to the output layer can be calculated in parallel).

In addition to this, the trained  DAEF models can be updated when new data arrives thanks to their incremental learning capacity. A node can add knowledge to its model without having to retrain from scratch, incorporating the new knowledge quickly and inexpensively. A DAEF network trained with a data partition can incorporate the knowledge obtained by a second   DAEF  network trained with a different partition if the latter shares the \textbf{U}$_{m_1}$ matrices of its encoder \cite{https://doi.org/10.1002/int.22296}, and the \textbf{M}$_{k}$, \textbf{U}$_{k}$, and \textbf{S}$_{k}$ matrices of each layer of its decoder \cite{DBLP:conf/iwann/Fontenla-Romero21}. By adding this information, the first DAEF network can recalculate its weights and will have learned incrementally.

If we are faced with an environment in which there are several nodes, such as an IoT scenario, where each node has a partition of the global dataset, we can take advantage of the incremental and distributed learning capacity of the DAEF network. Each node (device) would train a DAEF autoencoder network with its local data, and using a protocol such as MQTT, these nodes can publish their local model information through a broker to share their particular knowledge with the rest of the devices. The broker will be in charge of sending this information to the nodes that are subscribed to the updates, which will be able to aggregate the information received to their model. 

We consider the local dataset of each node as a partition of a global dataset, so all the nodes must use a DAEF autoencoder network with a similar architecture.  In order for the model information shared between nodes to be compatible with each other, the nodes must also use the same weights generated by the Xavier Glorot initialization scheme and the same bias. Before starting the training, one of the nodes must define the architecture, generate the weights and bias and publish them through the broker. {\bf Figure 3} shows this scenario using the MQTT protocol.

\begin{figure}
\centerline{\includegraphics[width=17.7pc]{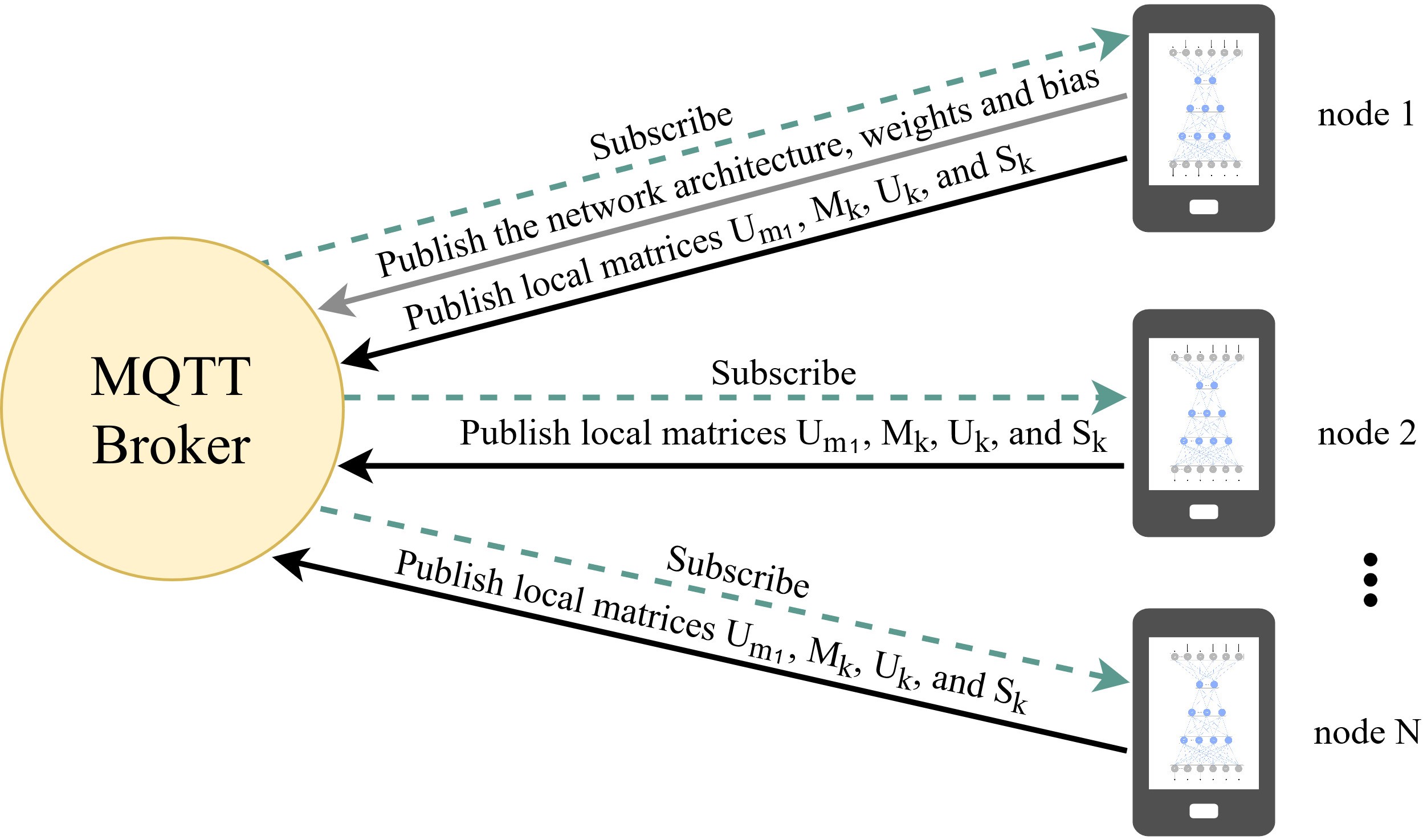}}
\caption{DAEF networks collaborating through an MQTT protocol.}
\label{figureMQTT}
\end{figure}

The private data of each node will be protected since the information that is sent through the broker to carry out the incremental learning is another. The data shared by each model will be the \textbf{U}$_{m_1}$ matrices of the encoder, and the \textbf{M}$_{k}$, \textbf{U}$_{k}$, and \textbf{S}$_{k}$ matrices of each layer of the decoder, from which the original data are not recoverable \cite{https://doi.org/10.1002/int.22296} \cite{DBLP:conf/iwann/Fontenla-Romero21}. The DAEF network matrices mentioned above are the only information needed to perform the federated learning, so if desired, the original dataset of each node can be removed to save space. Storing these matrices is not a problem since their size is independent of the number of instances of the original dataset.

Note that DAEF could also be used in a centralized scenario in which the information from the local models would be sent to a central node, which would be in charge of aggregating the information, obtaining the global model and sharing it with the network nodes.

\subsection{4.4 Pseudocode}\label{4_4}

Algorithm \ref{algorithm1} contains the pseudocode for the DAEF  training phase. The processes carried out in the encoder are described between lines 5 and 12. In line 7 the dimensionality of the data is reduced by means of SVD in a distributed way, obtaining the encoder weights and, in line 9, the encoder output. Between lines 13 and 19, the hidden layers of the decoder are trained one by one. For this, Algorithm \ref{algorithm2} is used (line 15). Between lines 20 and 25, the last layer of the decoder is trained directly using ROLANN.

\begin{algorithm}
    \caption{DAEF training phase}
    \label{algorithm1}
    \hspace*{\algorithmicindent} \textbf{Input:} $\mathbf{X}\in {\rm I\!R}^{m_0 \times n}$, training dataset ($m_0$ variables $\times$ $n$ samples); $a$, list of neurons per layer;  $\lambda_{H\!L}$ and $\lambda_{LL}$, regularization hyperparameters of the hidden and last layer; $f_{H\!L}$ and $f_{LL}$, activation functions of the hidden and last layers; $t$, available processes;\\
 	\hspace*{\algorithmicindent} \textbf{Output:} $M$, model composed of the weights and bias, the training output \textbf{H}$_{LL}$, \textbf{U}$_{1}$ and \textbf{S}$_{1}$ matrices of the encoder, the \textbf{M}$_{k}$, \textbf{U}$_{k}$, and \textbf{S}$_{k}$ matrices of each layer of the decoder, and the architecture;
    \begin{algorithmic}[1] 
		\Function{DAEF\_train}{}	
		\State $W_{list} = \varnothing $  \Comment{Layer weight list}
		\State $H_{list} = \varnothing $  \Comment{Layer output list}
		\State $matrices_{list} = \varnothing $  \Comment{Incremental learning matrices}
		\State $\mathbf{X}_{partitioned} = $ Split $ \mathbf{X} $ in $p$ partitions 
		\State $lat = a[1] $ \Comment{Latent space dimension}
		\State $\mathbf{U}_{1}, \mathbf{S}_{1} = DSV\!D(\mathbf{X}_{partitioned},lat) $ 
		\State $\mathbf{W}_{encoder} = \mathbf{U}_{1}$
		\State $\mathbf{H}_{encoder} = f_{H\!L}((\mathbf{W}_{encoder})^{T}\mathbf{X})$
		\State Append $\mathbf{W}_{encoder}$ to $W_{list}$ 
		\State Append $\mathbf{H}_{encoder}$ to $H_{list}$ 
		\State Append [$\mathbf{U}_{1}, \mathbf{S}_{1}]$ to $matrices_{list}$ 
        \State $hl_{decoder} =  length(a)-1$ \Comment{Decoder hidden layers}
        \For{\texttt{$l=2..hl_{decoder}$}} 
            \State  $ \mathbf{W}, \mathbf{b}, \mathbf{H}, matrices = TLD(\mathbf{H}_{list}[-1], a[l], \lambda_{H\!L}, f_{H\!L}, t)  $ 
            \State Append $\mathbf{W}$ to $W_{list}$ 
            \State Append $\mathbf{b}$ to $b_{list}$ 
            \State Append $\mathbf{H}$ to $H_{list}$ 
            \State Append $matrices$ to $matrices_{list}$ 
        \EndFor 
        \State $pool = $Pool$(t) $\Comment{Pool of $t$ processes}
		\State $\mathbf{W}_{LL}, \mathbf{b}_{LL}, \textbf{M}_{k}, \textbf{U}_{k}, \textbf{S}_{k} = pool.map(ROLAN\!N, (H_{list}[-1], \mathbf{X}, \lambda_{LL})) $ \Comment{Last layer ROLANN regularization in parallel}
		\State $\mathbf{H}_{LL} = f_{LL}((\mathbf{W}_{LL})^{T} H_{list}[-1])	$ 
		\State Append $\mathbf{W}_{LL}$ to $W_{list}$ 
		\State Append $\mathbf{b}_{LL}$ to $b_{list}$ 
		\State Append [$\textbf{M}_{k}$, $\textbf{U}_{k}$, $\textbf{S}_{k}$] to $matrices_{list}$ 
		\State $M = W_{list}, b_{list}, \mathbf{H}_{LL}, matrices_{list}, a$ 
        \State \Return $M$ 
		\EndFunction
    \end{algorithmic}
\end{algorithm}

Algorithm \ref{algorithm2} contains the pseudocode of the auxiliary function used in algorithm \ref{algorithm1} to train the different hidden layers of the decoder in a distributed way using an auxiliary autoencoder. In lines 2 and 3, the weights and bias are generated respectively, while in line 4 the output of the hidden layer is computed. Between lines 5 and 7, the decoder weights and the output are calculated using ROLANN in a distributed way. Since the weights with respect to each neuron of the output layer are calculated independently, the number of processes $t$ should not be higher.

\begin{algorithm}
    \caption{Train one layer of the decoder (TLD)}
    \label{algorithm2}
    \hspace*{\algorithmicindent} \textbf{Input:} $\mathbf{H}_l\in {\rm I\!R}^{m_l \times n}$, training data  from layer $l$ ($m_l$ variables $\times$ $n$ samples); $m_{l+1}$, number of neurons of the layer $l+1$;  $\lambda_{l+1}$, regularization hyperparameter of the hidden layer; $f_{l+1}$, activation function of the layer; $t$, available processes;\\
 	\hspace*{\algorithmicindent} \textbf{Output:} $\mathbf{W}_{l+1}$, weights of the layer $l+1$; $\mathbf{b}_{l+1}$ bias of the layer; $\mathbf{H}_{l+1}$, output of the layer $l+1$;
    \begin{algorithmic}[1] 
        \Function{TLD}{}	
		\State $\mathbf{W}_{c_{1}} = Xavier(m_l, m_{l+1}) $ \Comment{Initial weights}
		\State $\mathbf{b}_{c_{1}} = Random(m_{l+1},1) $ \Comment{Initial bias}
		\State $\mathbf{H}_{c_{1}} = f_{l+1}(\mathbf{W}^T_{c_{1}} \mathbf{H}_l + \mathbf{b}_{c_{1}}\mathbf{1}^T) $
		\State $pool = $Pool$(t) $\Comment{Pool of $t$ processes}
		\State $\mathbf{W}_{l+1}, \mathbf{b}_{l+1}, \textbf{M}_{k}, \textbf{U}_{k}, \textbf{S}_{k} = pool.map(ROLAN\!N, (\mathbf{H}_{c_1}, \mathbf{H}_l, \lambda_{l+1})) $ \Comment{ROLANN in parallel}
		\State $\mathbf{H}_{l+1} = f_{l+1}(\mathbf{W}_{l+1}^{T} \mathbf{H}_l + \mathbf{b}_{l+1}\mathbf{1}^T) $
        \State \Return $\mathbf{W}_{l+1}, \mathbf{b}_{l+1}, \mathbf{H}_{l+1}, [\textbf{M}_{k}, \textbf{U}_{k}, \textbf{S}_{k}]$
		\EndFunction
    \end{algorithmic}
\end{algorithm}

Algorithm \ref{algorithm3} contains the pseudocode for the DAEF prediction phase where the trained network will reconstruct a test sample. This algorithm can be useful for tasks such as anomaly detection.

\begin{algorithm}
    \caption{DAEF prediction phase}
    \label{algorithm3}
    \hspace*{\algorithmicindent} \textbf{Input:} $\mathbf{X}\in {\rm I\!R}^{m_0 \times n}$, test dataset ($m_0$ variables $\times$ $n$ samples); $W_{list}$, weights of the trained network; $b_{list}$, bias of the trained network; $f_{H\!L}$ and $f_{LL}$, activation functions of the hidden and last layers; $a$, list of neurons per layer;\\
 	\hspace*{\algorithmicindent} \textbf{Output:} $prediction$, reconstruction of the input $\mathbf{X}$ after passing through the network;
    \begin{algorithmic}[1] 
        \Function{DAEF\_predict}{}	
		\State $\mathbf{H} = f_{H\!L}((W_{list}[1])^{T}\mathbf{X})$
		\For{\texttt{$i=2..length(W_{list})-1$}} 
            \State $\mathbf{H} = f_{H\!L}((W_{list}[i])^{T}\mathbf{H} + b_{list}[i-1]\mathbf{1}^T)$
        \EndFor 
        \State $\mathbf{H} = f_{LL}((W_{list}[-1])^{T}\mathbf{H} + b_{list}[-1]\mathbf{1}^T)$
        \State \Return $\mathbf{H}$
		\EndFunction
    \end{algorithmic}
\end{algorithm}

\section{5. PRIVACY TREATMENT} \label{privacy}
In distributed environments (EC and FL), preserving the privacy of user data (nodes) is a critical aspect, even more so when they contain sensitive information such as personal data. Due to this, in this section we will analyze the privacy preservation capacity of the DAEF method. To do this we are going to consider two main threat scenarios \cite{privacy1}.

\subsection{5.1 Preventing direct leakage}\label{5_1}
In classic environments, it is common for the original data from the nodes to be sent to other nodes or to a central server, for example, for example, to be analyzed, preprocessed or to build a global model. This puts the privacy of the data at risk, which can be used maliciously and not to carry out the original tasks.

In the case of the DAEF method, the data shared to carry out the training of the global model is not the original data ($\mathbf{X}$). In the case of the encoder, each node $p$ computes an SVD using its local data ($\mathbf{X}_p$), and the information shared to carry out the federated learning is the product $\mathbf{U}_p \mathbf{S}_p$. Since the matrix $\mathbf{V}_p$ is neither calculated nor sent, the original data $\mathbf{X}_p$ cannot be retrieved through the factorization expression described in Equation 1. In the decoder, the federated learning is carried out using the $\mathbf{M}_p$, $\mathbf{U}_p$ and $\mathbf{S}_p$ matrices obtained through ROLANN regularization, so the original data is also kept safe.

Once the global model is trained, it is distributed to each of the local nodes $p$ to be used privately, so there is no direct data leakage in the operation phase.

\subsection{5.2 Preventing indirect leakage}\label{5_2}
Another possible scenario is one in which a malicious node impersonates a real participant of the distributed learning protocol to try to obtain the private data of other nodes. Due to the nature of their training, when we train iterative algorithms in a distributed way (such as traditional autoencoders), it is common for nodes to share their calculations and model parameters. In these cases, using this information and specific methods (inverse methods \cite{10.1145/2810103.2813677}, Generative Adversarial Networks \cite{10.1145/3133956.3134012}) the original data with which the training was carried out can be obtained, putting the privacy of the nodes at risk.

In the case of DAEF, the method is not iterative, so this type of attack is not a problem. The model parameters are calculated in a single step, so it is not possible to train GAN networks. In addition, as we have seen previously, stochastic gradients are not shared (they are not used) or sensitive information. In the articles taken as reference \cite{DBLP:conf/iwann/Fontenla-Romero21} \cite{https://doi.org/10.1002/int.22296} it has been shown that the original data cannot be recovered from the information sent by the node.
 
\section{6. RESULTS} \label{results}
In this section, several experiments are presented to show the behaviour of the proposed algorithm in real scenarios. Although autoencoder networks have several uses, the main task for which the DAEF method has been designed is anomaly detection. Given a trained DAEF network, the classification of new instances can be carried out by comparing their value at the network input and their value at the output. This is known as the reconstruction error, and since anomalies are very rare in these scenarios, instances corresponding to the normal class will have a low reconstruction error, while anomalies will emit a much higher. To do this, after training the network it will be necessary to establish an error threshold that allows classifying new data based on its reconstruction error. In this work we will define the threshold using the interquartile range (IQR) and also manually based on the percentage of anomalies existing in the dataset. To penalize higher errors, we will calculate the reconstruction errors using the MSE.

DAEF emerges as a fast alternative to perform anomaly detection in edge computing and federated learning environments. Iterative approaches achieve a high performance detecting anomalies, but their long training times make them unsuitable for these environments. The aim of this study is to check the performance achieved by DAEF compared to iterative deep autoencoders (AE).  Also, although by default DAEF uses Xavier Glorot initialization, other initializations such as totally random and orthogonal will be studied.

The algorithms have been evaluated over seven real datasets available in the UCI Machine Learning Repository and in the Kaggle website. The characteristics of these datasets are summarized in {\bf Table 1}. The data have been normalized using standard scalers with zero mean and unit variance. To assess the performance of each algorithm, the data has been split using a tenfold cross validation. The algorithms have been trained using only normal data, while the test phase included data from both classes (50\% normal and 50\% anomalies). The combinations of parameters chosen for each algorithm have been obtained by a grid search and are available in {\bf Appendix A}.

\begin{table}
\begin{center}
{\caption{Characteristics of the datasets used.}\label{tab:tabla1}}
\begin{tabular}{lccc}
\\[-6pt]
Dataset & Size & Anomalies & Dimension\\
\hline

Shuttle & 49097 & 3511 (7.2\%) & 9\\
Covertype & 286048 & 2747 (1.0\%) & 10\\
Pendigits & 6870 & 156 (2.3\%) & 16\\
Cardio & 1831 & 176 (9.6\%) & 21\\
Credit card & 284807 & 492 (0.2\%) & 29\\
Ionosphere & 351 & 126 (35.9\%) & 33\\
Optdigit & 5216 & 64 (2.9\%) & 62\\

\hline
\end{tabular}
\end{center}
\end{table}

The metric used to measure the performance of the algorithms was the F1-score, {\bf Table 2}  summarizes the mean test results. The chosen statistical test was Nemenyi, a non-parametric test which makes a pairwise comparison between models \cite{nemenyi2}.  Using a significance level of 5\% ($\alpha = 0.05 $) and the F1-scores obtained for each dataset independently, the best values in Table 2 have been highlighted in bold. As can be seen, the DAEF algorithm presents a robust behavior, achieving good performance for most datasets. The version of DAEF that uses the Xavier Glorot initialization stands out slightly from the others, matching the performance of the autoencoder in five of the seven datasets and surpassing it in another, according to the results of the statistical test. 

\begin{table}
\centering
{\caption{Average test F1-score $\pm$ standard deviation for the different datasets.}\label{tab:tabla2}}
\rule{0pt}{1ex}    
\resizebox{\columnwidth}{!}{%
\begin{tabular}{r|cccc}
\multicolumn{1}{r}{Dataset}
& \multicolumn{1}{c}{DAEF Ortho.}
& \multicolumn{1}{c}{DAEF Random}
& \multicolumn{1}{c}{DAEF Xavier}
& \multicolumn{1}{c}{AE} \\ \cline{1-5} 
Shuttle    & 95.0$\pm$0.6    &95.1$\pm$0.5     & 95.3$\pm$0.7     & \textbf{97.4$\pm$0.2}\\
Covertype    & \textbf{91.2$\pm$1.5}    &\textbf{90.5$\pm$1.5}     &\textbf{91.3$\pm$1.0}     & 85.7$\pm$3.4\\
Pendigits   & 73.9$\pm$10.4    &69.3$\pm$8.3     & \textbf{77.7$\pm$7.5}     & \textbf{85.9$\pm$2.6}\\
Cardio    & \textbf{87.5$\pm$1.2}   & \textbf{84.3$\pm$5.2}     & \textbf{87.1$\pm$3.6}     & \textbf{87.5$\pm$1.2} \\
Credit card    &\textbf{90.4$\pm$0.6}    & \textbf{90.6$\pm$0.4}     &\textbf{90.7$\pm$0.4}     & \textbf{90.5$\pm$0.8} \\
Ionosphere    &\textbf{90.5$\pm$5.1}    & \textbf{90.6$\pm$3.0}     &\textbf{89.5$\pm$8.3}     & \textbf{92.5$\pm$4.5} \\
Optdigit    & \textbf{72.0$\pm$5.1 }   & \textbf{73.4$\pm$8.5}    &\textbf{74.0$\pm$8.7 }    &\textbf{77.7$\pm$7.3 }\\
\hline
\end{tabular}%
}
\end{table}

Another statistical test was carried out to compare the global performance of the algorithms. The chosen test was again Nemenyi. Using a significance level of 5\% and the F1-scores of the algorithms for the different datasets, the three versions of DAEF and the autoencoder rank in the same position, represented graphically by {\bf Figure 4}. As can be seen, the null hypothesis that the algorithms obtain a similar performance is accepted, so we can affirm that in these tests DAEF obtained a similar performance to AE.

\begin{figure}
\centerline{\includegraphics[height=1.4in]{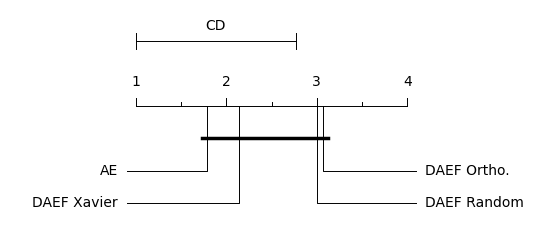}}
\caption{Graphical representation of Nemenyi test with $\alpha$ = 0.05. The critical distance (CD) obtained was 1.77.} \label{figure66}
\end{figure}

Because the execution of DAEF is parallelizable, the tests have been carried out using four cores. This was not possible with the autoencoder, which used a single core. {\bf Table 3} shows the mean training time of each algorithm (lower values than 0.05 have been represented as 0.0). Test times have not been included in this work because they are very low for all the algorithms. Due to DAEF's non-iterative training, its times are much shorter than those required by the traditional iterative autoencoder. The training times of DAEF have been between 15 and 68 times shorter in tests. Despite using a higher number of cores, the difference is significant. 

{\bf Table 4} shows an estimation of carbon dioxide emissions (grams of CO$_2$ emitted per kilowatt-hour) and power consumption (kWh) for the machine on which the tests were run \cite{codecarbon}. Since the three versions of DAEF obtained similar values, only the Xavier Glorot initialization has been included. As can be seen, both consumption and emissions are much lower compared to the traditional autoencoder, despite its parallel execution.

To compare the performance of DAEF against the reference method the experiments have been carried out in a traditional environment with a single machine. Despite this, we consider that the low computational cost of DAEF allows its use in an edge computing environment, characterized by large number of devices with less computing power. 

\begin{table}
\centering
{\caption{Average training time (seconds) $\pm$ standard deviation for the different datasets.}\label{tab:tabla3}}
\rule{0pt}{1ex}    
\resizebox{\columnwidth}{!}{%
\begin{tabular}{r|cccc}
\multicolumn{1}{r}{Dataset}
& \multicolumn{1}{c}{DAEF Ortho.}
& \multicolumn{1}{c}{DAEF Random}
& \multicolumn{1}{c}{DAEF Xavier}
& \multicolumn{1}{c}{AE} \\ \cline{1-5} 
Shuttle    & 2.1$\pm$0.1    &2.1$\pm$0.1     & 2.2$\pm$0.4     & 39.2$\pm$2.1\\
Covertype    & 4.8$\pm$0.4    &5.1$\pm$0.8     &4.7$\pm$0.2     & 341.3$\pm$7.6\\
Pendigits   & 2.2$\pm$0.0    &2.2$\pm$0.1     & 2.1$\pm$0.0     & 51.1$\pm$5.7\\
Cardio    &2.1$\pm$0.1   & 2.1$\pm$0.1     &1.9$\pm$0.6     &38.0$\pm$1.4 \\
Credit card    &58.4$\pm$1.2    & 58.9$\pm$1.0     &58.3$\pm$0.7     & 2249.1$\pm$18.2 \\
Ionosphere    &2.1$\pm$0.0    & 2.1$\pm$0.0     &2.1$\pm$0.0     & 30.6$\pm$3.2 \\
Optdigit    & 7.3$\pm$0.2    & 7.1$\pm$0.2    &7.3$\pm$0.2     & 125.3$\pm$4.9  \\
\hline
\end{tabular}%
}
\end{table}

\begin{table}
\centering
{\caption{Carbon dioxide emissions (grams CO$_2$/kWh) and power consumed (kWh) for the different datasets.}\label{tab:tabla3}}
\rule{0pt}{1ex}    
\resizebox{\columnwidth}{!}{%
\begin{tabular}{r|cccc}
\multicolumn{1}{r}{Dataset}
& \multicolumn{1}{c}{DAEF emissions}
& \multicolumn{1}{c}{DAEF power}
& \multicolumn{1}{c}{AE emissions}
& \multicolumn{1}{c}{AE power} \\ \cline{1-5} 
Shuttle    & $2.91\times 10^{-3}$  &$4.89\times 10^{-6}$    & $0.40$   & $6.72\times 10^{-4}$\\
Covertype  & $2.21\times 10^{-2}$  &$3.69\times 10^{-5} $   & $1.78$  & $2.98\times 10^{-3}$\\
Pendigits  & $1.38\times 10^{-2}$  &$2.31\times 10^{-5}  $  &$ 0.44 $  &$ 7.37\times 10^{-4}$\\
Cardio   & $3.02\times 10^{-3}$  &$5.07\times 10^{-6} $   & $0.39$   & $6.55\times 10^{-4}$\\
Credit card  &$ 0.32$ &$5.31\times 10^{-4}$    & $9.17$   & $1.54\times 10^{-2}$\\
Ionosphere    &$ 2.78\times 10^{-3} $ &$4.66\times 10^{-6}$    & $0.37$   & $6.16\times 10^{-4}$\\
Optdigit &$ 2.10\times 10^{-2}$  &$3.53\times 10^{-5}$    & $0.71$   &$ 1.19\times 10^{-3}$\\
\hline
\end{tabular}%
}
\end{table}

\section{CONCLUSION}

An alternative method to traditional deep autoencoder networks has been presented, with a robust performance in anomaly detection tests, and whose training time is much shorter than the reference method. Its distributed and incremental learning capacity, its low computational cost and its preservation of privacy make it a valid solution for edge computing and federated learning environments.

As future work, it would be interesting to test the algorithm in real edge computing or federated learning environments using different devices that act as independent nodes.

\newpage
\appendix
\section{Parameters used during training} \label{appendix1}

This appendix contains the values of the parameters finally chosen as the best for each method and dataset, listed in {\bf Table 5}.

The reconstruction error treshold ($\mu$) has been calculated using the IQR, where \textit{unusual} $ I\!Q\!R = Q_3 + 1.5 \times I\!Q\!R$, and \textit{extreme} $ I\!Q\!R = Q_3 + 3 \times I\!Q\!R $. 

\begin{table}
\centering
\begin{turn}{90}
\resizebox{8.7in}{!}{
\centering
\begin{tabular}{|c|c|c|c|c|}
\hline
Dataset & DAEF Ortho. & DAEF Random & DAEF Xavier & AE \\ 
\hline
Shuttle    & \begin{tabular}{c} Architecture: [9, 3, 5, 7, 9], \\ $\lambda_{H\!L}$: 0.7, $\lambda_{LL}$: 0.9, $\mu$: extreme IQR
  \end{tabular} & \begin{tabular}{c} Architecture: [9, 3, 5, 7, 9], \\ $\lambda_{H\!L}$: 0.7, $\lambda_{LL}$: 0.9, $\mu$: extreme IQR
  \end{tabular} & \begin{tabular}{c} Architecture: [9, 3, 5, 7, 9], \\ $\lambda_{H\!L}$: 0.8, $\lambda_{LL}$: 0.9, $\mu$: extreme IQR
  \end{tabular} &\begin{tabular}{c} Architecture: [9, 7, 5, 7, 9] \\ Epochs: 50, Contamination: 0.05  \end{tabular} \\
\hline
Covertype  & \begin{tabular}{c} Architecture:  [10, 2, 4, 6, 8, 10], \\ $\lambda_{H\!L}$: 0.7, $\lambda_{LL}$: 0.1, $\mu$: $Q_{90}$
  \end{tabular} & \begin{tabular}{c} Architecture:  [10, 2, 4, 6, 8, 10] , \\ $\lambda_{H\!L}$: 0.7, $\lambda_{LL}$: 0.1, $\mu$: $Q_{90}$
  \end{tabular} & \begin{tabular}{c} Architecture: [10, 2, 4, 6, 8, 10], \\ $\lambda_{H\!L}$: 0.7, $\lambda_{LL}$: 0.1, $\mu$: $Q_{90}$
  \end{tabular} &\begin{tabular}{c} Architecture: [10, 8, 6, 8, 10] \\ Epochs: 100, Contamination: 0.2  \end{tabular} \\
\hline
Pendigits   & \begin{tabular}{c} Architecture: [16, 4, 8, 12, 16], \\ $\lambda_{H\!L}$: 0.005, $\lambda_{LL}$: 0.5, $\mu$: extreme IQR
  \end{tabular} & \begin{tabular}{c} Architecture: [16, 8, 12, 16], \\ $\lambda_{H\!L}$: 0.005, $\lambda_{LL}$: 0.3, $\mu$: $Q_{90}$
  \end{tabular} & \begin{tabular}{c} Architecture: [16, 8, 12, 16], \\ $\lambda_{H\!L}$: 0.005, $\lambda_{LL}$: 0.7, $\mu$: $Q_{90}$
  \end{tabular} &\begin{tabular}{c} Architecture: [16, 12, 4, 12, 16] \\ Epochs: 100, Contamination: 0.2  \end{tabular} \\
\hline
Cardio  & \begin{tabular}{c} Architecture: [21, 4, 12, 21], \\ $\lambda_{H\!L}$: 0.9, $\lambda_{LL}$: 0.9, $\mu$: $ Q_{90}$
  \end{tabular} & \begin{tabular}{c} Architecture: [21, 4, 12, 21], \\ $\lambda_{H\!L}$: 0.9, $\lambda_{LL}$: 0.9, $\mu$: unusual IQR
  \end{tabular} & \begin{tabular}{c} Architecture: [21, 4, 8, 12, 16, 21], \\ $\lambda_{H\!L}$: 0.9, $\lambda_{LL}$: 0.9, $\mu$: $Q_{90}$
  \end{tabular} &\begin{tabular}{c} Architecture: [21, 12, 4, 12, 21] \\ Epochs: 100, Contamination: 0.2  \end{tabular} \\
\hline
Credit card  & \begin{tabular}{c} Architecture: [29, 15, 18, 21, 24, 27, 29] , \\ $\lambda_{H\!L}$: 0.005, $\lambda_{LL}$: 0.1, $\mu$: extreme IQR
  \end{tabular} & \begin{tabular}{c} Architecture: [29, 15, 18, 21, 24, 27, 29], \\ $\lambda_{H\!L}$: 0.8, $\lambda_{LL}$: 0.9, $\mu$: extreme IQR
  \end{tabular} & \begin{tabular}{c} Architecture: [29, 15, 18, 21, 24, 27, 29], \\ $\lambda_{H\!L}$: 0.8, $\lambda_{LL}$: 0.9, $\mu$: extreme IQR
  \end{tabular} &\begin{tabular}{c} Architecture: [29, 25, 20, 15, 20, 25, 29] \\ Epochs: 100, Contamination: 0.05  \end{tabular} \\
\hline
Ionosphere    & \begin{tabular}{c} Architecture: [33, 8, 14, 33], \\ $\lambda_{H\!L}$: 0.005, $\lambda_{LL}$: 0.7, $\mu$: extreme IQR
  \end{tabular} & \begin{tabular}{c} Architecture: [33, 8, 14, 33], \\ $\lambda_{H\!L}$: 0.1, $\lambda_{LL}$: 0.5, $\mu$: extreme IQR
  \end{tabular} & \begin{tabular}{c} Architecture: [33, 8, 14, 33], \\ $\lambda_{H\!L}$: 0.01, $\lambda_{LL}$: 0.8, $\mu$: extreme IQR
  \end{tabular} &\begin{tabular}{c} Architecture: [33, 25, 20, 15, 20, 25, 33]\\ Epochs: 100, Contamination: 0.1  \end{tabular} \\
\hline
Optdigit  & \begin{tabular}{c} Architecture: [62, 10, 20, 30, 40, 50, 62], \\ $\lambda_{H\!L}$: 0.005, $\lambda_{LL}$: 0.9, $\mu$: $Q_{80}$
  \end{tabular} & \begin{tabular}{c} Architecture: [62, 10, 20, 30, 40, 50, 62], \\ $\lambda_{H\!L}$: 0.9, $\lambda_{LL}$: 0.5, $\mu$: extreme IQR
  \end{tabular} & \begin{tabular}{c} Architecture: [62, 10, 20, 30, 40, 50, 62], \\ $\lambda_{H\!L}$: 0.8, $\lambda_{LL}$: 0.8, $\mu$: extreme IQR
  \end{tabular} &\begin{tabular}{c} Architecture: [62, 50, 40, 30, 20, 30, 40, 50, 62] \\ Epochs: 50, Contamination: 0.05  \end{tabular} \\
\hline
\end{tabular}
}
\end{turn}
\caption{Parameters used for training.}
\label{ta:dinos}
\end{table}

\section{ACKNOWLEDGMENT}

\section*{Acknowledgements}

This work was supported in part by grant \textit{Machine Learning on the Edge - Ayudas Fundaci\'on BBVA a Equipos de Investigaci\'on Cient\'ifica 2019}; the Spanish National Plan for Scientific and Technical Research and Innovation (PID2019-109238GB-C2); the Xunta de Galicia (ED431C 2018/34, ED431G 2019/01) and ERDF funds. CITIC is funded by Xunta de Galicia and ERDF funds.

\bibliographystyle{IEEEMicro}
\bibliography{references}

\begin{thebibliography}{10}

\bibitem{KHAN2019219}
Khan  W Z, Ahmed  E, Hakak  S, Yaqoob  I, and Ahmed  A, ``Edge computing: A
  survey,'' {\em Future Gener. Comput. Syst.}, vol.~97, pp.~219--235, 2019.

\bibitem{XIA2021100008}
Xia  Q, Ye  W, Tao  Z, Wu  J, and Li  Q, ``A survey of federated learning for
  edge computing: Research problems and solutions,'' {\em HCC}, vol.~1, no.~1,
  p.~100008, 2021.

\bibitem{10.1145/1541880.1541882}
Chandola  V, Banerjee  A, and Kumar  V, ``Anomaly detection: A survey,'' {\em
  CSUR}, vol.~41, jul 2009.

\bibitem{8744265}
Liu  S, Liu  L, Tang  J, Yu  B, Wang  Y, and Shi  W, ``Edge computing for
  autonomous driving: Opportunities and challenges,'' {\em Proceedings of the
  IEEE}, vol.~107, no.~8, pp.~1697--1716, 2019.

\bibitem{9139976}
Qiu  T, Chi  J, Zhou  X, Ning  Z, Atiquzzaman  M, and Wu  D O, ``Edge computing
  in industrial internet of things: Architecture, advances and challenges,''
  {\em IEEE Commun. Surv. Tutor.}, vol.~22, no.~4, pp.~2462--2488, 2020.

\bibitem{chandola}
Chandola  V, Banerjee  A, and Kumar  V, ``Anomaly detection: A survey,'' {\em
  CSUR}, vol.~41, no.~3, pp.~15:1--15:58, 2009.

\bibitem{khan}
Khan  S S and Madden  M G, ``One-class classification: Taxonomy of study and
  review of techniques,'' {\em Knowl}, vol.~abs/1312.0049, 2013.

\bibitem{10.5555/1756006.1953039}
Vincent  P, Larochelle  H, Lajoie  I, Bengio  Y, and Manzagol  P A, ``Stacked
  denoising autoencoders: Learning useful representations in a deep network
  with a local denoising criterion,'' {\em J. Mach. Learn. Res.}, vol.~11,
  p.~3371–3408, dec 2010.

\bibitem{nguyen2019diot}
Nguyen  T D, Marchal  S, Miettinen  M, Fereidooni  H, Asokan  N, and Sadeghi  A
  R, ``D\"iot: A federated self-learning anomaly detection system for {IoT},''
  2019.

\bibitem{8844663}
Hussain  B, Du  Q, Zhang  S, Imran  A, and Imran  M A, ``Mobile edge
  computing-based data-driven deep learning framework for anomaly detection,''
  {\em IEEE Access}, vol.~7, pp.~137656--137667, 2019.

\bibitem{sater2021federated}
Sater  R A and Hamza  A B, ``A federated learning approach to anomaly detection
  in smart buildings,'' 2021.

\bibitem{10.1145/3368926.3369705}
Zhao  Y, Chen  J, Wu  D, Teng  J, and Yu  S, ``Multi-task network anomaly
  detection using federated learning,'' in {\em SoICT 2019}, p.~273–279, ACM,
  2019.

\bibitem{app8122663}
Preuveneers  D, Rimmer  V, Tsingenopoulos  I, Spooren  J, Joosen  W, and Ilie
  Zudor  E, ``Chained anomaly detection models for federated learning: An
  intrusion detection case study,'' {\em Appl. Sci.}, vol.~8, no.~12, 2018.

\bibitem{8422402}
Luo  T and Nagarajan  S G, ``Distributed anomaly detection using autoencoder
  neural networks in {WSN} for {IoT},'' in {\em IEEE ICC}, pp.~1--6, 2018.

\bibitem{ngo2020adaptive}
Ngo  M V, Chaouchi  H, Luo  T, and Quek  T Q S, ``Adaptive anomaly detection
  for {IoT} data in hierarchical edge computing,'' 2020.

\bibitem{HUANG2006489}
Huang  G B, Zhu  Q Y, and Siew  C K, ``Extreme learning machine: Theory and
  applications,'' {\em Neurocomputing}, vol.~70, no.~1, pp.~489--501, 2006.
\newblock Neural Networks.

\bibitem{KasunRepresentationalLW}
Kasun  L, Zhou  H, Huang  G B, and Vong  C M, ``Representational learning with
  {ELMs} for {Big} {Data},'' {\em IEEE Intelligent Systems}, vol.~28,
  pp.~31--34, 11 2013.

\bibitem{DELM}
Ding  S, Zhang  N, Xu  X, Guo  L, and Zhang  J, ``Deep extreme learning machine
  and its application in {EEG} classification,'' {\em Math. Probl. Eng.},
  vol.~2015, pp.~1--11, 05 2015.

\bibitem{2021SSSS}
Ito  R, Tsukada  M, and Matsutani  H, ``An on-device federated learning
  approach for cooperative model update between edge devices,'' {\em IEEE
  Access}, vol.~9, p.~92986–92998, 2021.

\bibitem{4012031}
Liang  N y, Huang  G b, Saratchandran  P, and Sundararajan  N, ``A fast and
  accurate online sequential learning algorithm for feedforward networks,''
  {\em IEEE Transactions on Neural Networks}, vol.~17, no.~6, pp.~1411--1423,
  2006.

\bibitem{https://doi.org/10.1002/int.22296}
Fontenla Romero  O, Pérez Sánchez  B, and Guijarro{-}Berdi{\~{n}}as  B,
  ``{DSVD}-autoencoder: A scalable distributed privacy-preserving method for
  one-class classification,'' {\em Int. J. Intell. Syst.}, vol.~36, no.~1,
  pp.~177--199, 2021.

\bibitem{DBLP:conf/iwann/Fontenla-Romero21}
Fontenla{-}Romero  O, Guijarro{-}Berdi{\~{n}}as  B, and
  P{\'{e}}rez{-}S{\'{a}}nchez  B, ``Regularized one-layer neural networks for
  distributed and incremental environments,'' in {\em IWANN}, vol.~12862,
  pp.~343--355, Springer, 2021.

\bibitem{lannsvd}
Fontenla Romero  O, Pérez Sánchez  B, and Guijarro{-}Berdi{\~{n}}as  B,
  ``{LANN-SVD}: A non-iterative {SVD}-based learning algorithm for one-layer
  neural networks,'' {\em IEEE Trans. Neural Netw. Learn. Syst.}, vol.~29,
  pp.~3900--3905, 09 2017.

\bibitem{10.1016/j.patcog.2009.11.024}
Fontenla Romero  O, Guijarro{-}Berdi{\~{n}}as  B, P\'{e}rez S\'{a}nchez  B, and
  Alonso Betanzos  A, ``A new convex objective function for the supervised
  learning of single-layer neural networks,'' {\em Pattern Recogn.}, vol.~43,
  p.~1984–1992, may 2010.

\bibitem{RePEc:spr:psycho:v:1:y:1936:i:3:p:211-218}
Eckart  C and Young  G, ``The approximation of one matrix by another of lower
  rank,'' {\em Psychometrika}, vol.~1, no.~3, pp.~211--218, 1936.

\bibitem{2016dsvddistributed}
Iwen  M A and Ong  B W, ``A distributed and incremental {SVD} algorithm for
  agglomerative data analysis on large networks,'' {\em SIMAX}, vol.~37,
  p.~1699–1718, Jan 2016.

\bibitem{privacy1}
Shokri  R and Shmatikov  V, ``Privacy-preserving deep learning,'' in {\em
  Proceedings of the 22nd ACM SIGSAC Conference on Computer and Communications
  Security}, CCS '15, (New York, NY, USA), p.~1310–1321, Association for
  Computing Machinery, 2015.

\bibitem{10.1145/2810103.2813677}
Fredrikson  M, Jha  S, and Ristenpart  T, ``Model inversion attacks that
  exploit confidence information and basic countermeasures,'' in {\em
  Proceedings of the 22nd ACM SIGSAC Conference on Computer and Communications
  Security}, CCS '15, (New York, NY, USA), p.~1322–1333, Association for
  Computing Machinery, 2015.

\bibitem{10.1145/3133956.3134012}
Hitaj  B, Ateniese  G, and Perez Cruz  F, ``Deep models under the gan:
  Information leakage from collaborative deep learning,'' in {\em Proceedings
  of the 2017 ACM SIGSAC Conference on Computer and Communications Security},
  CCS '17, (New York, NY, USA), p.~603–618, Association for Computing
  Machinery, 2017.

\bibitem{nemenyi2}
Dem{\v{s}}ar  J, ``Statistical comparisons of classifiers over multiple data
  sets,'' {\em J. Mach. Learn. Res.}, vol.~7, no.~1, pp.~1--30, 2006.

\bibitem{codecarbon}
Schmidt  V, Goyal  K, Joshi  A, Feld  B, Conell  L, Laskaris  N, Blank  D,
  Wilson  J, Friedler  S, and Luccioni  S, ``{CodeCarbon: Estimate and Track
  Carbon Emissions from Machine Learning Computing},'' 2021.

\end{thebibliography}

\begin{IEEEbiography}{David Novoa-Paradela}{\,} (M) was born in Ourense, Spain, in 1996. He received his B.S. degree in computer science from the University of A Coruña in 2019, and his M.S. degree in artificial intelligence from the Menendez Pelayo International University in 2020. In October 2020 he started his Ph.D. thesis on the subject of "Machine Learning for Anomaly Detection: from surface to deep".
\end{IEEEbiography}

\begin{IEEEbiography}{Oscar Fontenla-Romero}{\,} (M) Ph.D. in Computer Science and Full Professor in Artificial Intelligence at the University of A Coruña. His research has focused on the development of new machine learning models, as well as its application in engineering and biomedicine areas. He has been part of the Board of Directors of the Spanish Association for Artificial Intelligence (AEPIA) from 2013 to 2018.  
\end{IEEEbiography}

\begin{IEEEbiography}{Bertha Guijarro-Berdiñas}{\,} (F) has a Ph.D. in Computer Science and is an Associate Professor at the University of A Coruña. Her research interests focus on Artificial Intelligence with special attention to the theoretical aspects of machine learning (distributed, online, scalable, sustainable and efficient learning, privacy preservation) and its applications. She has participated in more than 30 national and international projects, agreements with companies and is co-author of more than 100 articles. \end{IEEEbiography}

\end{document}